# Lamarck's Driving School: Discovering Autonomous Driving Training Strategies through Evolutionary Competition

Yichun Ye, He Zhang, Ye Tian*, *Senior Member, IEEE*, and Jian Sun, *Senior Member, IEEE*

***Abstract*—Autonomous driving capabilities depend strongly on the distribution of scenarios encountered during training. Existing methods commonly construct or dynamically adapt training scenario distributions using surrogate criteria such as realism, difficulty, or risk. However, these predefined surrogates may misrepresent training value, leading to inefficient use of training resources. To address this limitation, we propose a Lamarckian evolutionary framework that replaces surrogate-based guidance with competition among candidate distributions. We formulate training strategy discovery as a multi-stage bilevel optimization problem and use Lamarckian evolution algorithm to approximate its solution. At the outer level, Darwinian crossover, mutation, and selection explore the scenario distribution space; at the inner level, policy learning acquires new capabilities, and Lamarckian inheritance transfers them to subsequent stages, allowing scenario distributions and policy capabilities to co-evolve. The resulting evolutionary trajectories reveal recurring stage-wise regularities among high-value distributions, characterized by capability accumulation through stage-wise challenge rotation. We further distill these regularities into a lightweight, reusable Lamarckian Training Strategy. Experiments show that, compared with the baseline, the complete framework reduces performance loss by up to 25.07%, while the lightweight strategy still achieves a 19.13% reduction. These results demonstrate that evolutionary competition can both discover effective training strategies and reveal reusable stage-wise patterns in how the value of training distributions changes with policy capability. Code is available on GitHub.***

## I. INTRODUCTION

Learning-based autonomous driving has become a prominent research direction. Recent advances in end-to-end models [1-3], vision-language models [4-6], and world models [7-9] have further expanded the perception, reasoning, planning, and imagination capabilities of autonomous driving systems. As these models become more capable, further progress depends not only on model design, but also on the training scenarios from which they learn.

For learning-based autonomous driving, a driving scenario is a more natural unit of interaction than an isolated state-action pair [10]. A scenario specifies the road environment, traffic participants, interaction dynamics, environmental conditions, and task objectives [11]. Yet training gains are accumulated over a training set rather than a single scenario. The scenario distribution of that set determines what situations the policy encounters and how often it encounters them, thereby shaping the capabilities learned. Autonomous driving training therefore benefits not only from optimizing policy parameters under a given scenario distribution but also from optimizing the training scenario distribution itself. More broadly, research on embodied intelligence shows that environmental complexity can shape both acquired capabilities and learning efficiency [12]. This motivates us to study scenario distributions as the environmental structure of autonomous driving training.

Existing studies design or dynamically adapt training scenario distributions according to surrogate criteria. Naturalistic scenario replay emphasizes realism by sampling from recorded traffic data [13-15]; curriculum learning organizes or adapts scenarios according to difficulty, learning progress, or model weaknesses [16-21]; and safety-critical scenario generation prioritizes risk, failure, or criticality [22-24]. These strategies share an implicit assumption: the better a distribution scores under the chosen surrogate criterion, the greater its training value.

However, distributions favored by these criteria do not necessarily produce better post-training performance. The underlying problem is an objective mismatch: the signal used to design the training distribution differs from the outcome that training is intended to improve [25-27]. Consequently, surrogate-driven design can lead to inefficient use of training resources and obscure reusable patterns in which distributions are valuable at different stages of policy development.

To align distribution design with post-training performance, we let candidate distributions compete according to the performance they produce after policy training rather than assigning value through prescribed surrogate rules. Repeated competition produces high-value training paths and evolutionary records from which reusable training patterns can be discovered.

*Research supported by the National Outstanding Youth Science Fund Project of National Natural Science Foundation of China under Grant No. [52422215] and the National Engineering Research Center of Road Safety Control Technology under Grant number [2025GCZXKFKT19A].

Yichun Ye, Ye Tian, and Jian Sun are with the Department of Traffic Engineering and the Key Laboratory of Road and Traffic Engineering, Ministry of Education, Tongji University, Shanghai, China (e-mail: 2610276@tongji.edu.cn, tianye@tongji.edu.cn, sunjian@tongji.edu.cn). He Zhang is with the School of Transportation and Logistics, Southwest Jiaotong University, Chengdu, Sichuan, China (e-mail: zhanghe@swjtu.edu.cn). Corresponding author: Ye Tian (tianye@tongji.edu.cn).

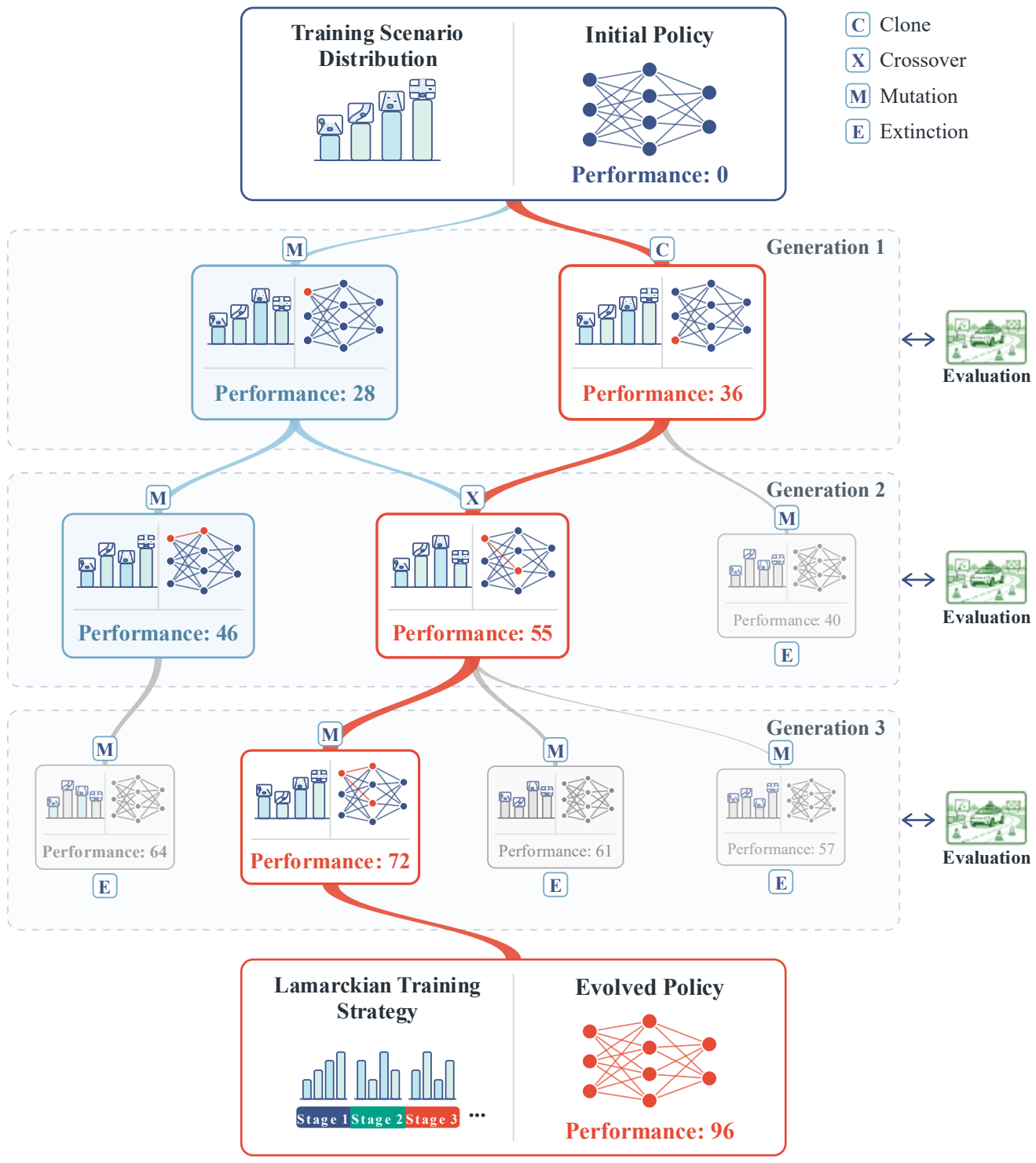

Fig. 1. Overview of the proposed framework, in which scenario distributions and driving policies evolve through competition based on post-training evaluation and capability inheritance

Fig. 1 summarizes how scenario distributions and policy capabilities evolve through competition and inheritance. The value of a candidate distribution becomes observable only after policy training and cannot be directly differentiated with respect to the distribution parameters, making strategy discovery a multi-stage black-box optimization problem. We formulate this search as a multi-stage bilevel problem and approximate it through Lamarckian evolution. At each stage, Darwinian crossover, mutation, and selection [28] explore the scenario distribution space, while the policy is trained separately on each candidate distribution. Post-training performance determines which individuals survive, and Lamarckian inheritance [29] passes the acquired policy capabilities to the next stage. The resulting evolutionary record reveals which distributions are valuable at successive capability stages without prescribing in advance how the training distribution should change.

We use complete Lamarckian evolution as a large-scale discovery experiment and analyze the high-value distributions retained across generations. The resulting trajectories reveal an empirical pattern of capability accumulation through stage-wise challenge rotation. Because repeating complete evolution is computationally expensive, we distill these regularities into a reusable, lightweight Lamarckian Training Strategy (LTS) that preserves the discovered training logic without further evolutionary search.

The contributions of this paper are summarized as follows.

- The first formulation of training strategy discovery for autonomous driving as a multi-stage bilevel problem driven by post-training performance rather than surrogate criteria.
- A Lamarckian evolutionary framework that enables training strategies to emerge through distribution competition and capability inheritance.
- Evolutionary analysis reveals an interpretable pattern of capability accumulation through stage-wise challenge rotation, which is distilled into a lightweight and reusable Lamarckian Training Strategy.
- Experiments demonstrate the effectiveness of the Lamarckian Training Strategy, which outperforms the Replay baseline and all comparison methods in overall score.

The rest of this paper is organized as follows. Section II reviews related work. Section III formulates autonomous driving training as a bilevel optimization problem over multiple stages. Section IV presents the proposed Lamarckian evolutionary framework. Section V reports the evolutionary search, strategy validation, and ablation studies. Section VI concludes the paper and discusses future directions.

## II. RELATED WORK

Existing studies on training scenarios can be broadly grouped into three paradigms: naturalistic scenario replay, curriculum-based scenario organization, and safety-critical scenario generation. These paradigms organize training scenarios around realism, difficulty, or risk, but generally select distributions using surrogate criteria rather than post-training performance.

### *A. Naturalistic Scenario Replay*

Naturalistic scenario replay provides a foundational source of training scenarios for autonomous driving. Datasets such as Waymo Open Dataset [13], nuScenes [14], and ApolloScape [15] cover real road structures, traffic participant behaviors, and diverse driving tasks, supporting perception, prediction, planning, and end-to-end driving model training. Real-world replay preserves natural traffic distributions, but naturalistic data is dominated by common low risk cases, providing insufficient exposure to long-tail interactions and safety-critical situations [30]. Although scenario augmentation, domain randomization, and benchmark design improve coverage, the training scenario distribution is still mostly treated as an externally given condition rather than an optimizable strategy variable.

### *B. Curriculum-Based Methods*

Curriculum-based methods treat training content selection as a higher-level decision. Existing approaches use manual curricula, fixed rules, automated sampling, policy-dependent environment generation, regret, or learning potential to design, prioritize, or adapt scenarios online [16], [18], [31-37]. Population-based training also combines evolutionary competition with parameter inheritance, but optimizes learning hyperparameters rather than training data or environment distributions [38]. Inspired by evidence that environmental complexity shapes agent capabilities and learnability [12], we use complete Lamarckian evolution to let scenario distributions compete, identify recurring patterns among elite distributions, and distill them into a reusable training strategy rather than another online adaptation rule.

### C. Safety-Critical Scenario Generation

Safety-critical scenario generation focuses on long-tail risks and system vulnerabilities. Related methods construct high risk scenarios through optimization, reinforcement learning, generative models, or large models to expose failure modes under boundary conditions [24, 39]. For example, Closed-loop Adversarial Training (CAT) formulates closed-loop training as a joint process of driving policy learning and adversarial environment generation, and continuously trains the policy in safety-critical scenarios [23]. These methods increase exposure to challenging scenarios, but they usually optimize collision rate, failure rate, risk exposure, or criticality. Such objectives may bias training toward worst-case scenarios without revealing which distribution sequences produce reusable improvements in task-level performance.

## III. PROBLEM DEFINITION

### A. Learning Formulation for Autonomous Driving

In autonomous driving learning, the basic training unit is not an isolated state-action pair, but a complete closed-loop driving scenario. A scenario with horizon $T$ is denoted in (1).

$$X_T = (s_0, a_0, s_1, a_1, \dots, s_{T-1}, a_{T-1}, s_T) \tag{1}$$

Here, $s_t$ denotes the traffic state at time t, including road structure, weather, and the states of traffic participants. $a_t$ denotes the action taken by the ego vehicle under state $s_t$.

Let $q$ denote the training scenario distribution, which specifies the scenario factors sampled during training. Although closed-loop trajectories are jointly induced by $q$ and the ego policy $\theta$, we write $X_T \sim q$ for simplicity.

Under a given $q$, autonomous driving learning optimizes the policy parameter $\theta$ with respect to a training objective $J$. Here, $J$ may represent the expected cumulative reward in reinforcement learning or the negative loss in imitation or supervised learning. (2) gives a typical form based on cumulative reward.

$$J(\theta, q) = \mathbb{E}_{X_T \sim q}\left[\sum_{t=0}^{T-1} \gamma^t \, r(s_t, a_t, s_{t+1})\right] \tag{2}$$

The standard learning problem is written in (3).

$$\theta^*(q) = \underset{\theta}{\operatorname{argmax}}\, J(\theta, q) \tag{3}$$

This formulation captures the conventional setting, in which policy parameters are optimized under a given training scenario distribution. Naturalistic scenario replay, curriculum-based training, and adversarial training mainly differ in how $q$ is constructed, rather than in the basic learning form itself.

If different training scenario distributions lead to different learning outcomes, $q$ should be treated not merely as an externally specified condition but as an optimization variable in the training strategy.

### B. Training Strategy Optimization Problem

The training objective and the final evaluation objective are not fully aligned. The training stage directly optimizes $J(\theta, q)$, such as cumulative reward or training loss, whereas final driving performance is measured by a comprehensive evaluation function $F$. Since $F$ is usually an aggregated metric and cannot be simply decomposed into rewards at individual time steps, optimizing $J$ under a given $q$ does not necessarily optimize F, as shown in (4).

$$\underset{\theta}{\operatorname{argmax}} J(\theta, q) \nRightarrow \underset{\theta}{\operatorname{argmax}} F(\theta) \tag{4}$$

This mismatch indicates that the training scenario distribution affects not only the training process but also the final evaluation performance. We therefore treat the sequence of training scenario distributions as an optimizable training strategy. Let q denote the sequence of distributions used over finite-step stages $K$ . The resulting multi-stage bilevel optimization problem is written in (5).

$$\begin{cases} q^* = \underset{q \in Q}{\arg\max}\, F(\theta_K) \\ \theta_{k+1} = U(\theta_k, q_k), 0 \le k < K \end{cases} \tag{5}$$

Here, $Q$ is the candidate space of training scenario distributions, and U denotes the finite-step policy update under a candidate distribution. Direct sequence-level optimization is intractable because evaluating a candidate requires policy training. We therefore approximate it stage by stage and define the stage-k training value as the post-training evaluation performance in (6). This stage-wise formulation is a greedy approximation to (5), rather than an exact decomposition or a guarantee of global optimality.

$$V_k(q_k; \theta_k) = F(U(\theta_k, q_k)) \tag{6}$$

Here, $V_k(q_k; \theta_k)$ is the post-training value of $q_k$ under the current policy capability $\theta_k$ and the fixed stage-wise training budget. It is conditional on the inherited policy state and the evaluation protocol, rather than an intrinsic or policy-independent property of $q_k$. The outer-level selection problem at stage k is therefore written as (7).

$$q_k^* = \underset{q_k \in Q}{\arg\max}\, V_k(q_k; \theta_k) \tag{7}$$

This formulation uses post-training evaluation performance to define stage-conditioned distribution value, which becomes the outer-level fitness signal in the Lamarckian evolution framework of Section IV.

## IV. METHODOLOGY

Section III shows that the training scenario distribution should not be treated merely as an externally given condition, but as the core object of training strategy optimization. Based on this problem definition, this paper proposes a Lamarckian evolutionary framework for automatically discovering high-value scenario distributions during training. The framework couples scenario search with policy learning: the outer level continuously generates and selects high-value training scenario distributions, while the inner level updates policy capability under these distributions and transfers the learned policy state to the next stage. Fig. 2 summarizes the problem formulation and the resulting framework.

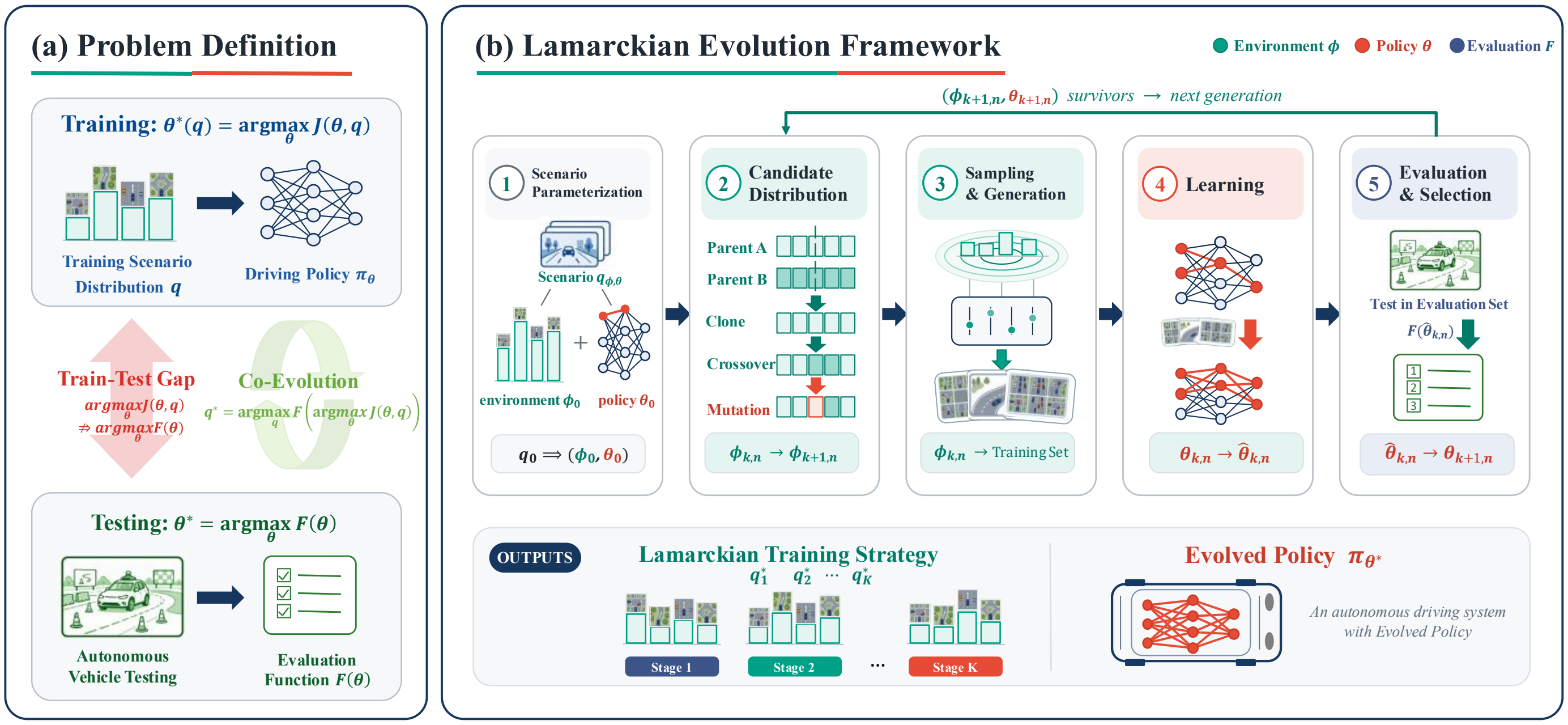

Fig. 2. Overview of the problem formulation and Lamarckian evolutionary framework.

The method consists of scenario parameterization in Section IV-A, Lamarckian evolution in Section IV-B, scenario generation and sampling in Section IV-C, and distillation into a reusable Lamarckian Training Strategy in Section IV-D.

*A. Scenario Parameterization*

To make the optimization operational, we decompose closed-loop trajectory generation. The training scenario distribution q specifies the initial traffic states, environment dynamics, and background behavior. Together with the ego policy, it induces the closed-loop trajectory distribution. For a trajectory $X_T$ with horizon $T$:

$$
\begin{aligned}
P(X_T) &= P(s_0, a_0, s_1, a_1, \dots, s_{T-1}, a_{T-1}, s_T) \\
&= P(s_0) \prod_{t=0}^{T-1} P(s_{t+1}|s_t, a_t) \cdot \pi_\theta(a_t|s_t) \\
&= P_\phi(s_0) \prod_{t=0}^{T-1} P_\phi(s_{t+1}|s_t, a_t, b_t) P_\phi(b_t|s_t) \pi_\theta(a_t|s_t) \,.
\end{aligned}
\tag{8}
$$

Here, $P_\phi(s_0)$ is the initial scenario distribution, $P_\phi(s_{t+1}|s_t, a_t, b_t)$ is the environment dynamics, $P_\phi(b_t|s_t)$ is the background-vehicle policy, and $\pi_\theta(a_t|s_t)$ is the ego policy.

Eq. (8) separates environment parameter $\phi$, which characterizes initial states, background behavior, and dynamics, from policy parameter $\theta$, which characterizes the ego policy's capability. Thus:

$$X_T \sim q_{\theta,\phi}. \tag{9}$$

At stage k, selecting a distribution reduces to searching over environment parameters. Let $q(\phi_k)$ denote the scenario distribution induced by $\phi_k$. The stage-wise problem derived from (7) is:

$$\phi_k^* = \arg\max_{\phi_k \in \Phi} V_k(q_{\phi_k}; \theta_k) \tag{10}$$

Here, $V_k(q_{\phi_k}; \theta_k)$ is the conditional post-training value produced by the distribution induced by $\phi_k^*$, and the selected distribution is $q_k^* = q(\phi_k^*)$. This value is defined under the current inherited policy state, fixed training budget, and evaluation protocol.

Thus, $\theta$ is updated through learning at the inner level, whereas $\phi$ is updated through evolution at the outer level. $\phi$ contains parameters governing initial states, environment dynamics, and background policies, turning abstract distribution optimization into tractable parameter search. The environment parameters are listed in *Table A*.1.

*B. Lamarckian Evolution Framework*

After parameterization, each stage becomes a finite-step bilevel process: the outer level searches over environment parameters, and the inner level updates the inherited policy under the induced distribution.

Following standard fitness-based selection in evolutionary computation [40], we use evaluation performance after a fixed training budget as fitness. This directly matches the operational objective and avoids favoring a poorly performing inherited policy merely because it admits a large numerical gain, as expressed in (11).

$$f_{k,n} = F(U(\theta_{k,n}, q_{\phi_{k,n}})) \tag{11}$$

For candidate n at stage k, the distribution induced by $\phi_{k,n}$ is used by U for the finite-step update from the inherited policy state. Candidate distributions compete through the post-training performance realized within their inherited policy lineages.

The challenge is twofold: the environment-parameter space is high-dimensional and heterogeneous, and each candidate's fitness becomes observable only after policy training and evaluation. We approximate this black-box discovery process using Lamarckian evolution across multiple environments.

In Lamarckian evolution, individuals are evaluated after learning, and the acquired policy capability state is preserved. Here, environment parameters undergo evolutionary operations, while policy parameters are updated through finite-step training and capability inheritance. The framework consists of five steps; the sampling procedure, evaluation function, and algorithm parameters are detailed in **Algorithm 1**, *Table A*.3, and *Table A*.4, respectively.

1) Population

A complete Lamarckian evolution run consists of multiple training stages, and each stage contains several candidate individuals. The n-th candidate individual at stage k is composed of an environment parameter and a policy parameter, as shown in (12).

$$\mathcal{P}_k = \{(\phi_{k,n}, \theta_{k,n})\}_{n=1}^{N} \quad (12)$$

Here, the environment parameter determines the corresponding training scenario distribution, while the policy parameter represents the policy capability inherited by the individual at the current stage. In the initial stage, all candidate individuals start from the same initial policy parameter and initialize their environment parameters based on scenario statistics from naturalistic driving data. In later stages, candidate individuals are generated jointly through parent inheritance, crossover, mutation, and injection of high-fitness environment parameters.

2) Crossover and Mutation

Crossover combines existing high-fitness environment parameters so that different scenario attributes, such as road structure, traffic state, and behavioral statistics, can be recombined. Mutation perturbs local environment parameters to explore new training scenario distributions. The generation of new candidate environment parameters is expressed in (13).

$$\phi_{k,c} = \mathrm{Mut}\Big(\mathrm{Cross}\big(\phi_{k-1,p1}, \phi_{k-1,p2}\big)\Big) \quad (13)$$

Crossover and mutation operate only on environment parameters and do not directly modify policy parameters. This separation keeps the method logically consistent: training content is organized through evolution at the outer level, whereas policy capability is acquired through policy training at the inner level.

3) Learning

For each candidate environment parameter, the corresponding training scenario set is first sampled from the distribution it induces. The policy is then initialized with that individual's inherited policy parameter and undergoes finite-step training, as shown in (14).

$$\hat{\theta}_{k,n} = U\big(\theta_{k,n}, q_{\phi_{k,n}}\big) \quad (14)$$

The updated policy parameter $\hat{\theta}_{k,n}$ represents the updated capability state acquired under the candidate training scenario distribution. Candidate individuals must be evaluated and selected before they can enter the next stage.

4) Evaluation

After finite-step policy training, the unified evaluation function $F$ scores the updated policy associated with each candidate individual, thereby measuring the training utility of the candidate environment parameter at the current capability stage, as shown in (15).

$$F\big(\hat{\theta}_{k,n}\big) = \beta_0 + \sum \beta_i \cdot M_i\big(\hat{\theta}_{k,n}\big) \quad (15)$$

The evaluation function F reflects overall performance under a unified protocol rather than training reward or loss. Selection therefore depends on evaluation performance after training, not surrogate criteria such as realism, difficulty, or risk. The metrics and weights are provided in *Table A*.3. During evolutionary search, F is computed on the fitness evaluation set, while the final test set is reserved for final performance reporting.

5) Selection

High-fitness individuals are selected as parents. Let $\mathcal{I}_k$ be their index set; individuals in the next stage inherit the capability states acquired through finite-step training from their selected parents, as in (16). Environment parameters enter the next stage through inheritance, crossover, mutation, or the injection of high-fitness environment parameters.

$$\begin{cases} \mathcal{I}_k = \mathrm{Select}\big(\{F(\hat{\theta}_{k,n})\}_{n=1}^{N}\big) \\ \theta_{k+1,i} = \hat{\theta}_{k,p_i}, \qquad p_i \in \mathcal{I}_k \end{cases} \quad (16)$$

What is preserved is not only the high-value training scenario distribution represented by the environment parameter, but also the policy capability state acquired through finite-step training under that distribution. Thus, policy capability can accumulate continuously along the evolutionary path, reflecting the core Lamarckian feature of inheritance of acquired characteristics.

### *C. Scenario Generation and Sampling*

The environment parameter provides a parameterized representation of the training scenario distribution, but actual training requires converting it into an executable training scenario set. To this end, we adopt a pipeline that constructs a scenario pool, samples scenarios to match target distributions, and checks feasibility, thereby mapping abstract environment parameters to executable training data.

We construct a unified scenario pool from a common set of 400 Waymo source scenarios and two sets of variants derived from these same sources.

1) The naturalistic component consists of the 400 Waymo source scenarios and preserves their road structures, traffic-participant trajectories, and interaction patterns.
2) Generalized variants are generated by TrafficGen [41] exclusively from these 400 source scenarios to expand the coverage of vehicle layouts, initial states, and trajectory distributions.
3) Safety-critical variants are generated by CAT [23] exclusively from the same 400 source scenarios to supplement long-tail and safety-critical training cases.

Given a target environment parameter, the sampler constructs a training scenario set from the unified pool such that its empirical statistics match the target distribution as closely as possible. The sampling objective combines the matching of feature moments and binned distributions. Specifically, the sampler replaces one scenario at a time to minimize the weighted discrepancy between the target distribution and the empirical distribution of the current scenario set, and a replacement is accepted only when it reduces the matching error. The sampling procedure is detailed in **Algorithm 1**.

When the scenario pool sufficiently covers the target distribution, the training scenario set is constructed through sampling and recombination. Otherwise, TrafficGen or CAT is invoked on the same Waymo source scenarios to supply missing attributes. TrafficGen is used when greater diversity in road structure, traffic flow, or motion patterns is required, whereas CAT is used when lower TTC or THW values are required. No method has access to additional independent real-world source scenarios.

During crossover and mutation, candidate environment parameters may violate physical laws or task logic. To keep the search within a valid scenario space, we introduce the GeneLock mechanism. GeneLock is not designed to increase scenario difficulty; rather, it constrains evolutionary search to the feasible region of the scenario space. It checks physical constraints, constraints on task feasibility, and operational design domain (ODD) constraints. Candidates that violate these constraints are rejected or projected back to the feasible domain.

### *D. High-Value Distributions and Training Strategy Distillation*

Complete Lamarckian evolution can discover high-value training scenario distributions, but directly using it as a routine training strategy is computationally expensive. It evaluates multiple candidate distributions at each stage, and the resulting evolutionary path may depend on the algorithm, initialization, and training noise. Thus, repeating complete Lamarckian evolution whenever a new policy is trained would limit the practical usability of the method.

To address this issue, we treat complete Lamarckian evolution as a discovery process for high-value training scenario distributions rather than as the final reusable training strategy. The reusable output is a stable Lamarckian Training Strategy with fixed training stages and low computational cost. We therefore design a mechanism that extracts high-value scenario distributions and distills them into a training strategy, which compresses the environment parameters discovered during evolution into a fixed sequence of training scenario distributions for each stage.

The first distillation step records the environment-parameter vectors of high-fitness individuals after each evolutionary stage. These vectors are ranked by post-training performance and deduplicated to form the elite distribution log, which supports both elite reuse during evolution and subsequent strategy distillation.

Training strategy distillation does not train a new control policy network. In the second step, generation-wise normalized feature means of the elite distributions are examined for sustained shifts across multiple features between adjacent generations. In this experiment, these change points define six contiguous stages; the resulting segmentation is descriptive and is not claimed to be unique. In the third step, the parameter vectors of all elite distributions within each stage are averaged to define the target distribution, after which GeneLock projects the averaged vector back to the feasible domain when necessary. A fixed-size training scenario set is then sampled for each stage.

The Lamarckian Training Strategy (LTS) can be used for policy training without repeating complete evolution. With 12 generations, 10 candidates per generation, and 25k training steps per candidate, complete evolution requires 3.0 million policy-training steps, excluding evaluation, whereas one LTS run uses 300k steps. This one-time discovery cost can be amortized when the fixed strategy is reused for subsequent policy training. Accordingly, all claims of improved training efficiency refer to the distilled LTS, whereas complete evolution is evaluated primarily by the quality of the discovered path and distributions. The experiments evaluate LTS in terms of learning efficiency, final performance, and performance on held-out scenarios from both datasets.

## V. EXPERIMENTS

This section evaluates the proposed framework through three experiments. First, complete Lamarckian evolution is used as a discovery experiment to identify progressively improving paths and valuable scenario-distribution patterns. Second, the distilled Lamarckian Training Strategy (LTS) is compared with Replay [13], Curriculum [16], and CAT [23] under the same policy-training budget to determine whether the discovered patterns provide efficient, reusable training without online evolution and improve performance on the disjoint Waymo and nuScenes test scenarios. Third, the ablation study removes scenario search and capability inheritance separately to examine their roles in the observed behavior.

### *A. Experimental Setup*

The experiments are implemented in the MetaDrive simulator [42], and TD3 [43] is used as the policy-learning algorithm. All real-world source scenarios used for training originate from 400 Waymo scenarios. TrafficGen and CAT are applied only to these sources to derive 1,510 generalized variants and 658 safety-critical variants; these outputs do not introduce additional independent real-world source data. Replay, Curriculum, CAT, and the proposed method therefore share the same source scenarios and maximum training budget, while their actual training sequences differ according to their organization or generation mechanisms. The fitness evaluation set, which is disjoint from both the training scenarios and the final test sets, contains 50 Waymo and 50 nuScenes scenarios, whereas the final test sets contain disjoint sets of 100 Waymo and 83 nuScenes scenarios. Because the nuScenes evaluation scenarios participate in fitness evaluation, the nuScenes test results measure performance on held-out scenarios from a second dataset rather than zero-shot generalization to an entirely unseen domain. The evaluation metrics cover efficiency, safety, traffic compliance, and performance across the two datasets and are combined by a weighted evaluation function. Additional details are provided in *Table A.*5.

### *B. Lamarckian Evolution Experiment*

Complete Lamarckian evolution is used to discover a high-performing training path and stage-wise patterns in scenario distributions for subsequent strategy distillation. Best Evolution denotes the path with the highest fitness selected according to post-training performance on the fitness evaluation set before the final test set is accessed.

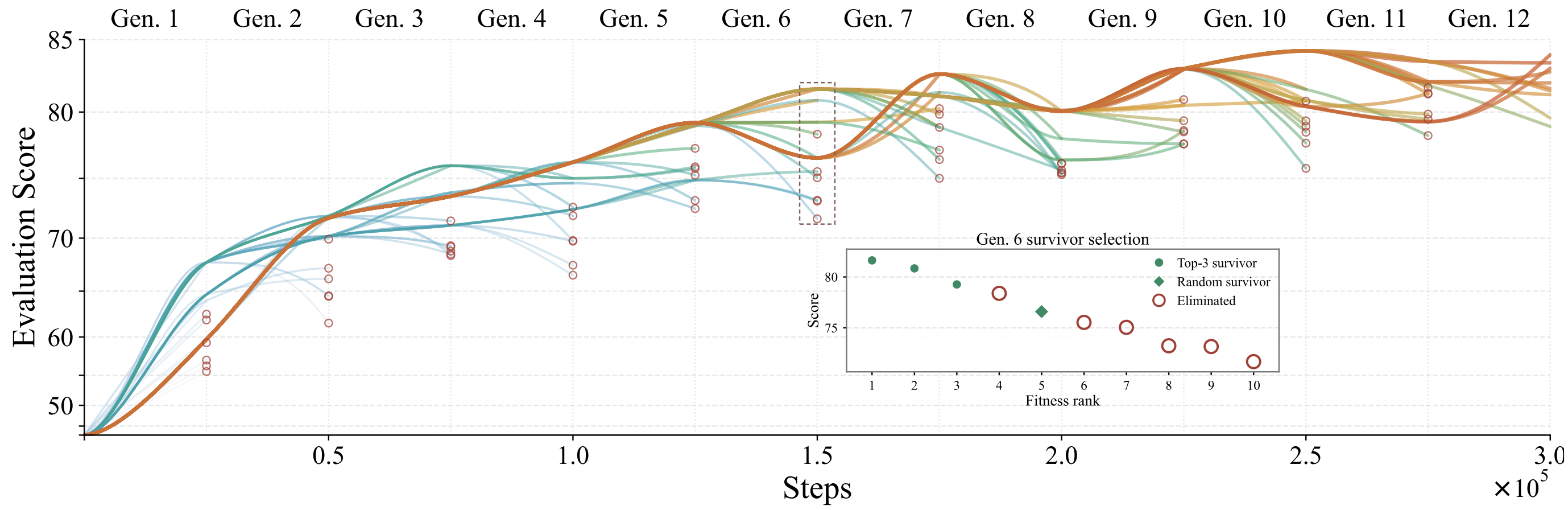

Fig. 3. Evolutionary tree produced by Lamarckian evolution.

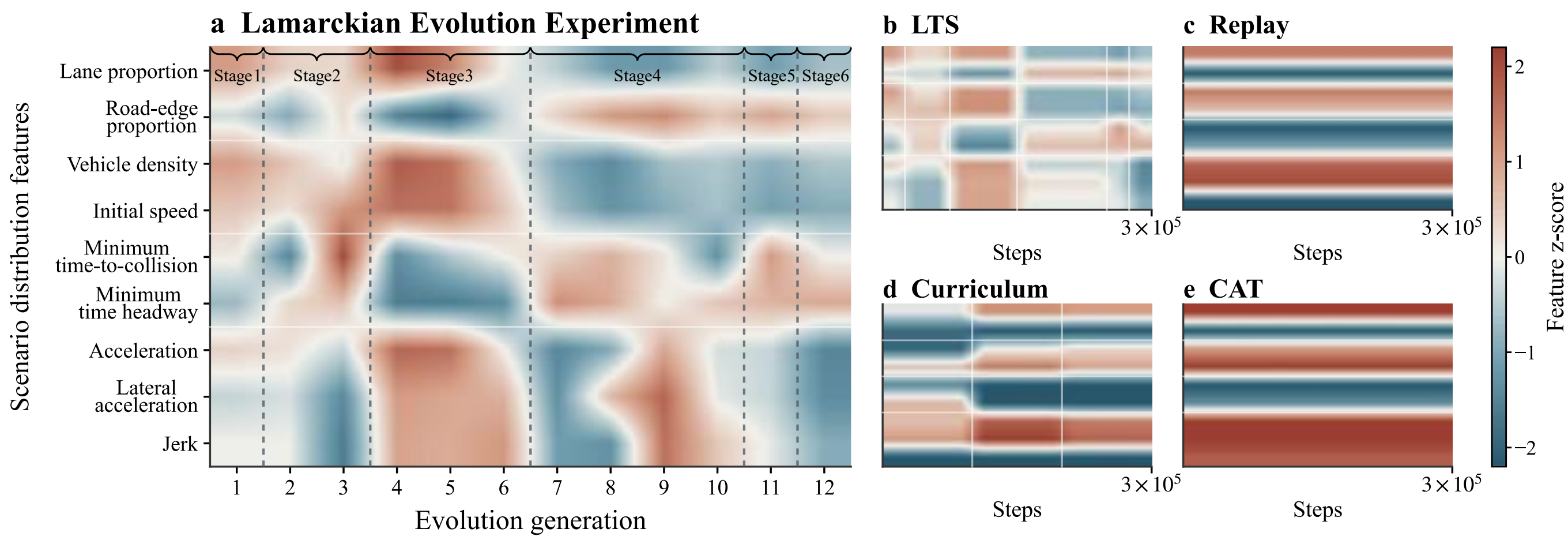

Fig. 4. Stage-wise scenario distribution characteristics of different methods.

The search comprises 12 generations with a population size of 10. In each generation, every candidate undergoes 25,000 training steps on its scenario distribution and is then evaluated under the same protocol. Each generation contains four cloned, three crossover, and three mutation candidates; an elite gene pool reuses historically high-value scenario distributions to improve search efficiency.

Fig. 3 shows the evolutionary tree obtained from 12 generations with a population size of 10. Each curve represents an evolutionary branch, and a branch ends when its individual is eliminated. The branches diverge substantially in evaluation performance, while selection retains individuals with higher scores and produces several progressively improving paths.

Fig. 4 compares nine scenario distribution features across five methods. Lane and road-edge proportions measure their respective shares of map features. Higher lane proportion suggests a regular, lane-dominated layout, whereas higher road-edge proportion indicates stronger boundaries and greater geometric complexity. Vehicle density and initial speed describe traffic intensity and the initial motion state. Minimum time to collision (TTC) and time headway (THW) measure conflict urgency and car-following gaps, indicating safety margins. Acceleration, lateral acceleration, and jerk describe motion intensity, abruptness, and control smoothness. The heatmaps report feature values standardized as z-scores.

Fig. 4(a) visualizes the elite distribution statistics recorded during complete Lamarckian evolution. Their dominant features shift across generations, and sustained changes in the normalized feature means allow the evolutionary record to be descriptively divided into six contiguous stages, with the learning focus rotating among basic driving, smooth control, traffic interaction, road structure, safety margins, and policy stability. Fig. 4(b) presents LTS, whose target distribution in each stage is obtained by averaging the elite parameter vectors assigned to that stage. The close correspondence between Fig. 4(a) and (b) therefore reflects the distillation procedure. Fig. 4(c)-(e) show Replay, Curriculum, and CAT, respectively. Replay maintains the naturalistic Waymo distribution, Curriculum follows a conventional three-stage progression from easy to difficult scenarios, and CAT emphasizes more adversarial and hazardous scenarios.

Overall, the evolved strategy follows neither a monotonic easy-to-hard curriculum nor a continual increase in risk. Instead, its dominant learning challenge shifts across stages, while Lamarckian inheritance preserves previously acquired capabilities. This pattern is characterized as capability accumulation through stage-wise challenge rotation. Detailed definitions and distribution characteristics of the six stages are provided in Table 1.

Table 1. Stage-wise regularities of the Lamarckian Training Strategy.

| Stage | Steps | Focus | Scenario Distribution Characteristics |
|---|---|---|---|
| Stage1 | 0-25k | Basic Driving Learning | Lane proportion and vehicle density are higher, while THW is lower. This creates dense car-following traffic that emphasizes lane keeping and speed regulation. |
| Stage2 | 25-75k | Smooth Control Learning | Lateral acceleration and jerk are lower, while THW is higher. This creates smooth, well-spaced traffic that emphasizes steering smoothness and car-following gap control. |
| Stage3 | 75-150k | Traffic Interaction Learning | Vehicle density, initial speed, acceleration, and jerk are higher, while THW is lower. This creates fast, dense, dynamic traffic that emphasizes yielding, merging, and coordination. |
| Stage4 | 150-250k | Road Structure Learning | Road-edge proportion is higher, while lane proportion, density, and initial speed are lower. This creates sparse traffic that emphasizes adaptation to road layout and boundaries. |
| Stage5 | 250-275k | Safety Margin Learning | TTC, THW, and road-edge proportion are higher, while initial speed is lower. This creates slow, well-spaced traffic that emphasizes early avoidance and safe spacing. |
| Stage6 | 275-300k | Policy Stability Learning | Acceleration, lateral acceleration, and jerk are lower, while THW is higher. This creates stable, low-disturbance traffic that emphasizes behavioral consistency and capability retention. |

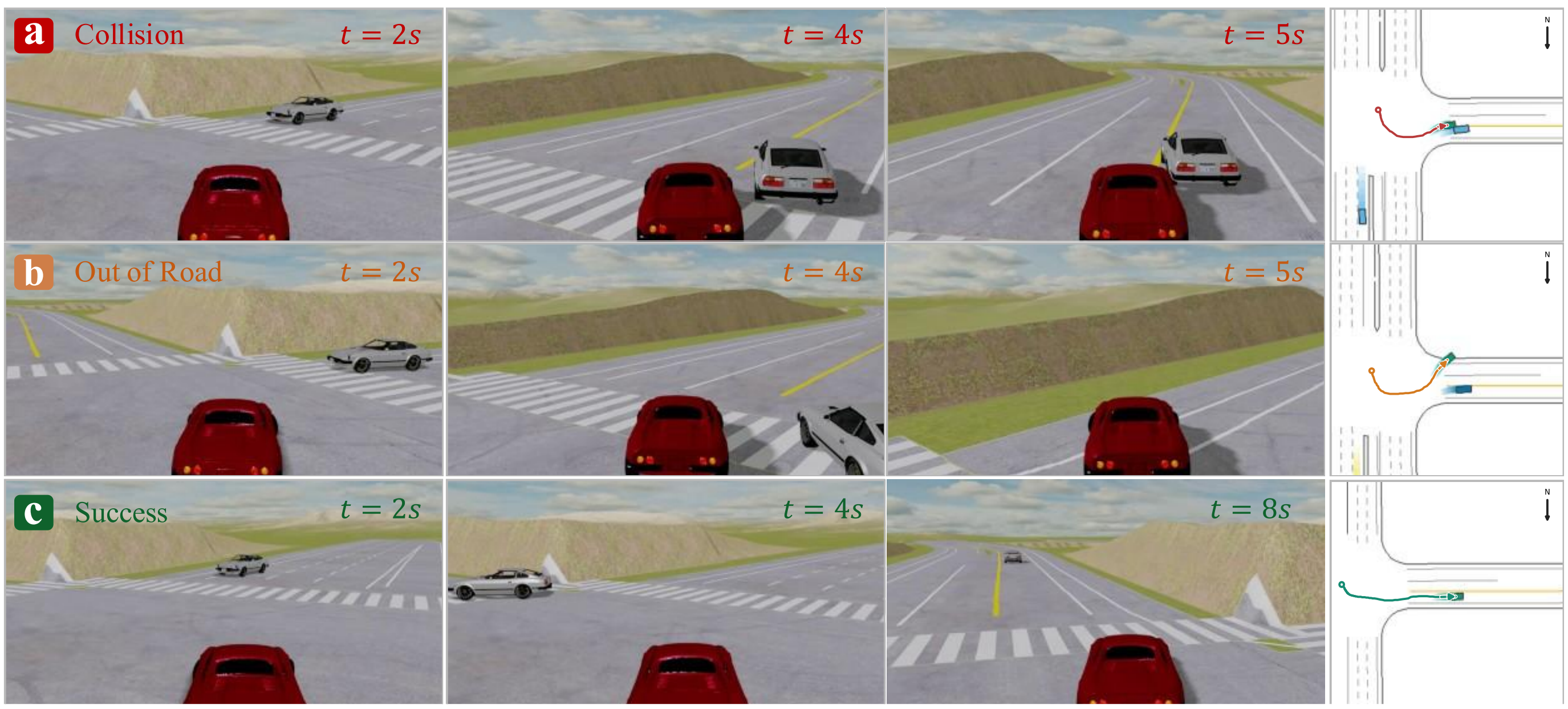

Fig. 5. Capability divergence after one generation of evolution. Starting from the same initial policy, different candidate training scenario distributions are associated with distinct post-training driving behaviors.

Fig. 5 compares three policies trained for one generation from the same initial policy. In the same scenario, they produce three outcomes: collision, out-of-road failure, and successful completion. Their training distributions differ in interaction intensity, motion variability, and safety margins, which may contribute to the observed behavioral divergence.

In Fig. 5(a), the candidate training scenario distribution has the highest interaction intensity among the three cases: a vehicle density of 10.17 veh/km, an initial speed of 3.61 m/s, longitudinal and lateral accelerations of 1.71 and 0.50 m/s², and a jerk of 1.15 m/s³. It also has the smallest safety margins, with minimum TTC and THW values of 6.88 and 2.66 s. The resulting policy accepts a narrow gap and collides with a surrounding vehicle. This outcome may be associated with frequent exposure to dense, dynamic interactions and limited yielding margins.

In Fig. 5(b), the candidate training scenario distribution has the lowest vehicle density but retains noticeable motion variability, with an initial speed of 2.25 m/s, longitudinal and lateral accelerations of 0.85 and 0.28 m/s², and a jerk of 0.50 m/s³. The resulting policy avoids the immediate conflict but subsequently leaves the road after a large lateral correction. One possible explanation is that this distribution provides experience in conflict response but less coverage of lateral recovery near road boundaries.

In Fig. 5(c), the candidate training scenario distribution has the smoothest dynamic profile: its vehicle density is 7.29 veh/km; its initial speed, longitudinal acceleration, lateral acceleration, and jerk are the lowest at 1.44 m/s, 0.60 m/s², 0.21 m/s², and 0.33 m/s³; and its minimum TTC and THW are the largest at 7.82 and 3.91 s. The resulting policy slows earlier, steers smoothly, and completes the scenario. These cases do not establish a causal effect for any single scenario factor, but they show that different training scenario distributions can yield substantially different outcomes after training under the same initial policy and evaluation scenario. This observation provides an intuitive illustration of why the elite distributions analyzed in Fig. 4 matter for training.

Table 2. Results of different training strategies.

| Method | Score↑ | Waymo | | | nuScenes | | |
|---|---|---|---|---|---|---|---|
| | | Success % ↑ | Collision % ↓ | Out-of-Road %↓ | Success %↑ | Collision %↓ | Out-of-Road % ↓ |
| *Replay* | 78.62 ± 1.55 | 79.60 ± 1.14 | 18.20 ± 1.10 | 2.80 ± 1.10 | 65.54 ± 4.64 | 22.17 ± 2.64 | 9.40 ± 1.98 |
| *Curriculum* | 79.77 ± 1.36 | 81.00 ± 2.92 | 15.20 ± 1.30 | 2.60 ± 2.30 | 66.51 ± 1.32 | 22.89 ± 3.71 | 9.16 ± 3.96 |
| *CAT* | 80.09 ± 1.87 | 81.60 ± 1.95 | 16.20 ± 1.92 | 2.00 ± 1.22 | 66.51 ± 3.94 | 20.00 ± 2.50 | 10.60 ± 2.61 |
| **LTS** | **82.71 ± 0.42** | **85.40 ± 2.30** | **13.20 ± 3.11** | **1.60 ± 0.89** | **69.16 ± 2.90** | **21.93 ± 4.92** | **5.30 ± 1.83** |
| **Best Evolution** | **83.98** | **82.00** | **16.00** | **2.00** | **78.31** | **13.25** | **7.23** |

Based on the elite-distribution patterns in Fig. 4(a), the high-fitness scenario distributions discovered during complete evolution are distilled into a lightweight LTS with a fixed sequence of stages. This process does not copy the single path with the highest score. Instead, it records elite distribution vectors, uses their generation-wise descriptive shifts to group contiguous generations into six stages, and averages the elite vectors within each stage to obtain the corresponding LTS target distribution. The resulting strategy is summarized in Table 1, while the exact stage vectors, sampled scenario sets, and policy checkpoints are available in the project repository.

### *C. Validation of the Lamarckian Training Strategy*

We compare the Lamarckian Training Strategy (LTS) with Replay, Curriculum, and CAT under the same learning algorithm, maximum training budget, evaluation protocol, and common set of 400 Waymo source scenarios. Each method organizes or generates its actual training sequence exclusively from these common sources according to its own mechanism. Replay is the reference baseline, whereas Curriculum and CAT are comparison methods. For each training strategy, five policies are independently trained from scratch using different random seeds. Table 2 reports mean and standard deviation across these runs, and the shaded regions in Fig. 6 denote 95% confidence intervals. Performance is evaluated every 25k steps, giving all methods the same twelve evaluation points within the 300k-step budget.

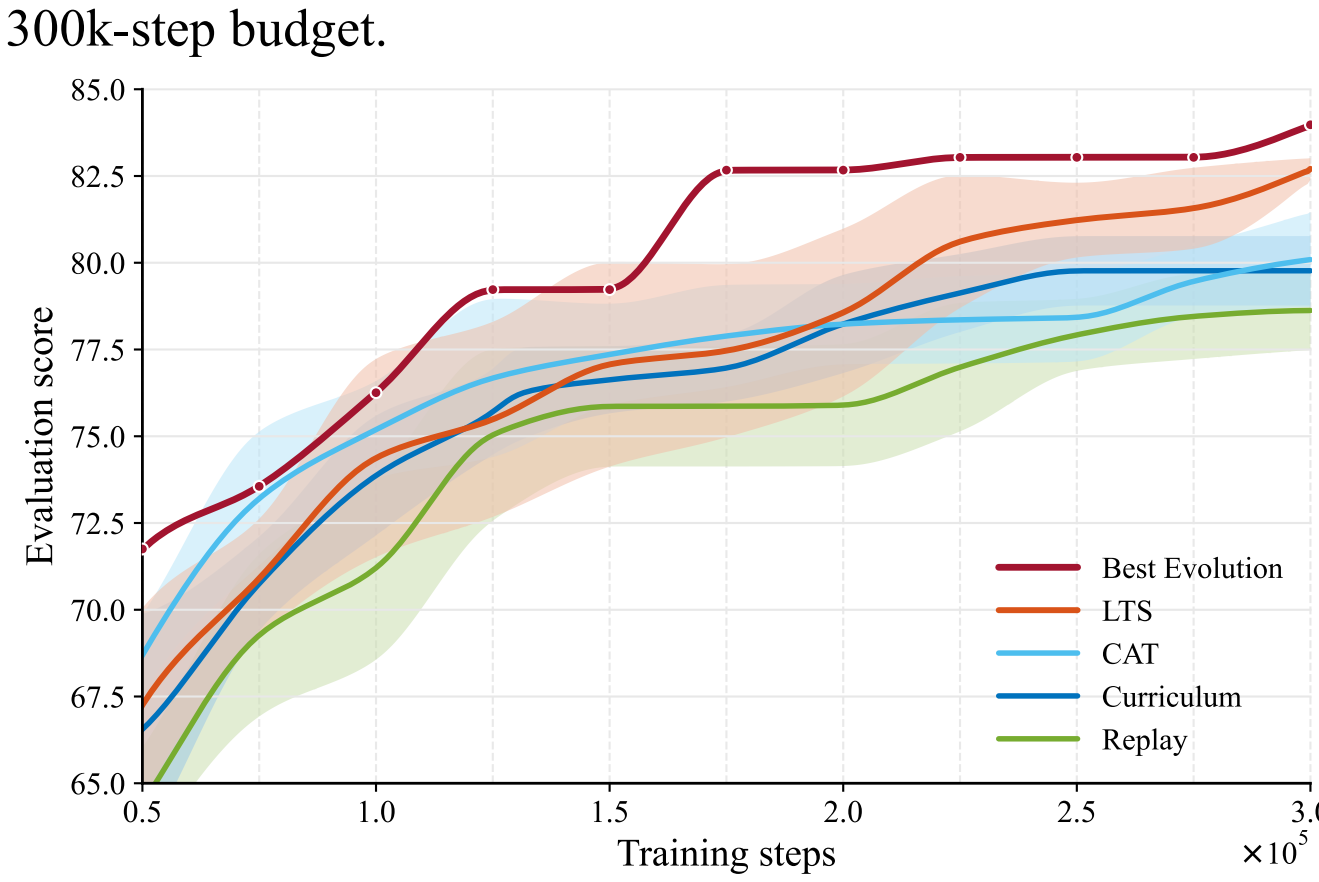

Fig. 6. Best-so-far performance curves and confidence intervals for different training strategies.

Fig. 6 and Table 2 show that LTS achieves the highest mean overall score in the validation experiment, outperforming Replay, Curriculum, and CAT. On the Waymo test set, it reaches a success rate of 85.40%, with collision and out-of-road (OOR) rates of 13.20% and 1.60%, respectively. On the nuScenes test set, its success rate reaches 69.16%, and its OOR rate is 5.30%. With all component metrics expressed in percentage points, the evaluation function in *Table A*.3 computes a weighted overall score in [0, 100] from the test results, and performance loss is defined as 100 minus this score. Relative to Replay, the corresponding loss reductions are 25.07% for Best Evolution and 19.13% for LTS.

Replay performs reasonably well on the Waymo test set but less strongly on the held-out nuScenes scenarios. Curriculum and CAT are guided by difficulty and safety-criticality surrogates, respectively. LTS instead rotates the dominant training challenge across stages, moving from traffic adaptation and control calibration to interaction learning, road-structure adaptation, safety-margin reinforcement, and policy consolidation. The results therefore support stage-specific distribution organization rather than reliance on a single surrogate criterion.

Overall, complete Lamarckian evolution reveals more than a high-scoring path tied to the original search: it also uncovers reusable patterns in training scenario distributions. Although the Lamarckian Training Strategy does not reach the score achieved by Best Evolution, it outperforms the Replay baseline and all compared methods without online evolution, suggesting that evolutionary analysis can be converted into a lightweight training strategy.

### *D. Ablation Study*

The ablation study uses the same computational budget across stages, evaluation sets, evaluation function, initial policy, and random seed as the full evolutionary experiment. The w/o Outer Search variant replaces environment-parameter inheritance, crossover, mutation, and elite-pool selection with random sampling at each stage from the same scenario pool, while retaining policy-state inheritance across stages. The w/o Capability Inheritance variant retains the evolved distributions selected for each stage but deliberately reinitializes every stage from the fixed initial policy before the 25k-step update.

Table 3. Ablation results for the key mechanisms.

| Setting | Score ↑ | Δ | Waymo | | | nuScenes | | |
|---|---|---|---|---|---|---|---|---|
| | | | Success % ↑ | Collision % ↓ | Out-of-Road %↓ | Success %↑ | Collision %↓ | Out-of-Road % ↓ |
| **Full Model** | **83.98** | — | **82.00** | **16.00** | **2.00** | **78.31** | **13.25** | **7.23** |
| w/o Search | 71.78 | -12.20 | 70.00 | 17.00 | 11.00 | 56.63 | 25.30 | 15.66 |
| w/o Inheritance | 67.89 | -16.09 | 67.00 | 25.00 | 8.00 | 50.60 | 25.30 | 22.89 |

The w/o Outer Search variant examines whether evolutionary search identifies useful stage-specific distributions. The w/o Capability Inheritance variant is a deliberately destructive intervention that examines whether capabilities must be preserved across stages: although total computation over all stages is unchanged, no individual policy accumulates more than one stage of learning. It is therefore interpreted as a mechanistic check rather than a policy-age-matched estimate of the isolated numerical effect of inheritance.

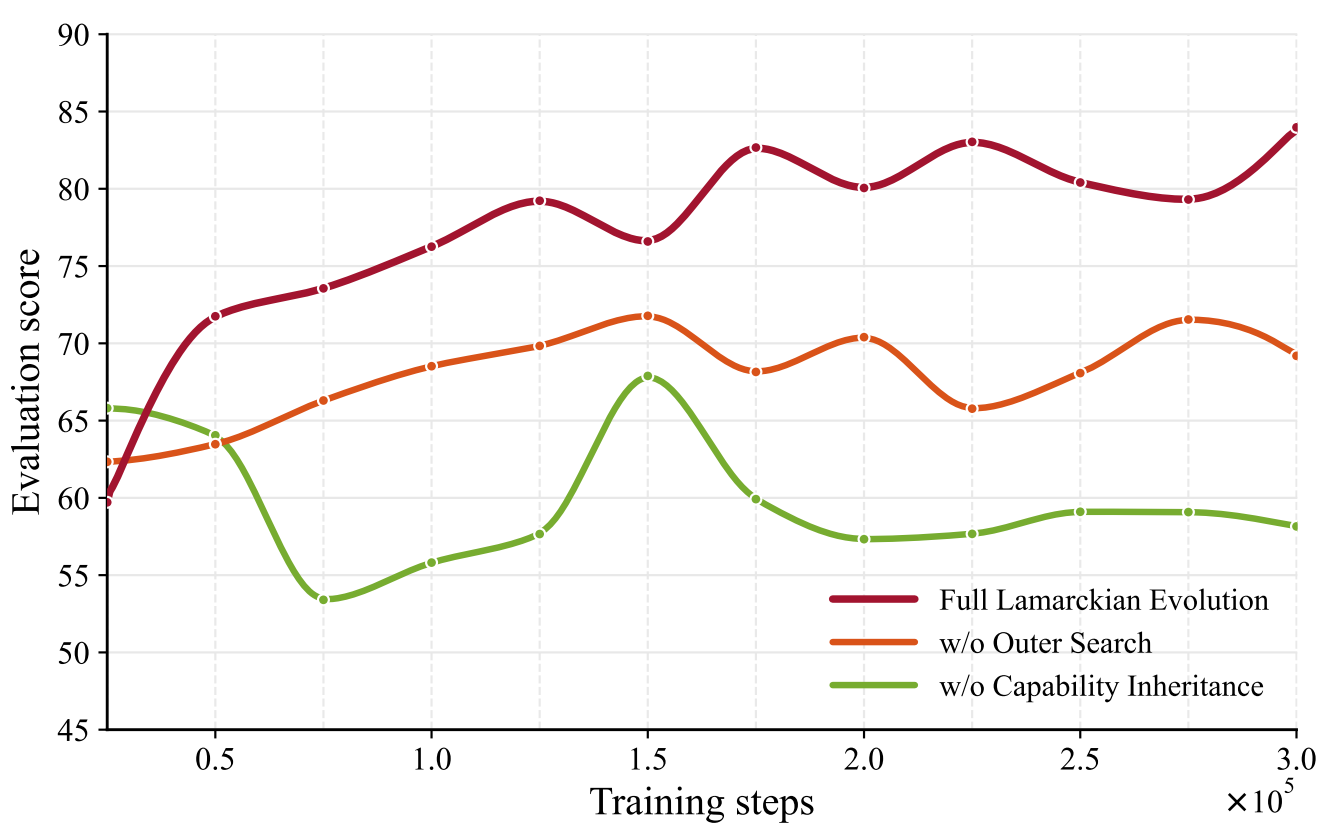

Fig. 7. Learning curves for the ablation study.

Table 3 and Fig. 7 report the ablation results. Removing outer search reduces the best score from 83.98 to 71.78 and the held-out nuScenes success rate from 78.31% to 56.63%, indicating that random scenario selection does not reproduce the stage-specific distributions identified by evolutionary search. Removing capability inheritance further reduces the best score to 67.89 and the held-out nuScenes success rate to 50.60%. Because every stage then restarts from the initial policy, capabilities acquired in earlier stages are repeatedly discarded. This degradation provides a mechanistic sanity check that high-quality stage distributions alone are insufficient without cross-stage capability accumulation; its magnitude should not be interpreted as an isolated causal effect size for inheritance.

### *E. Discussion*

The experiments highlight two roles of the Lamarckian evolutionary framework. First, complete evolution is a discovery process that identifies high-performing paths and valuable distribution patterns across policy-development stages. Second, those patterns can be distilled into a low-cost LTS that is used without online population search. This study is intended as a proof-of-concept investigation of strategy discovery through evolutionary competition under a controlled simulator protocol, rather than an engineering study of a production-scale online training system. The emphasis is therefore on whether competitive evolution can reveal reusable distribution patterns and whether those patterns remain effective after distillation.

## VI. CONCLUSION

This paper proposes a Lamarckian evolutionary framework for optimizing training strategies in autonomous driving, treating training scenario distributions as searchable optimization variables. At the outer level, evolution organizes the training scenario distributions; at the inner level, learning updates policy capability; and Lamarckian inheritance transfers the learned policy state across stages. Together, these mechanisms form a closed loop of distribution search, policy learning, performance-based competition, and capability inheritance.

The results show that complete Lamarckian evolution identifies high-value scenario distributions for individual stages that can be distilled into an effective Lamarckian Training Strategy with six stages. Under the same training budget, the Lamarckian Training Strategy achieves higher overall performance than Replay, Curriculum, and CAT and remains stable across repeated runs. Best Evolution illustrates the additional potential of online evolutionary search, while the ablations show that both search over scenario distributions and capability inheritance contribute to the observed gains.

This work has several limitations. The experiments are conducted primarily in MetaDrive with TD3, and transfer to other learning algorithms, simulators, and large real-world scenario libraries remains to be verified. Complete Lamarckian evolution is also computationally expensive. Best Evolution is reported as a reference from a single search run, and the quality of the distilled Lamarckian Training Strategy depends on the scenario generator and evaluation protocol.

Future work will evaluate transfer across additional learning paradigms and algorithms for autonomous driving, incorporate richer generative models and real-world road testing scenarios, and investigate more efficient evolutionary search and strategy distillation. Further theoretical analysis and high-fidelity closed-loop validation will also be pursued.

## APPENDIX

*Table A.1. Scenario Parameterization*

| Component of $\phi$ | Interpretation | Concrete Parameter |
|---|---|---|
| $P_\phi(s_0)$ | Initial scenario distribution | road_ratio_lane, road_ratio_road_line, road_ratio_road_edge, lane_count, lane_width, curvature, veh_density_init, speed_init, distance_init |
| $P_\phi(s_{t+1}\|s_t, a_t, b_t)$ | Environment dynamics | max_engine_force, max_brake_force, wheel_friction, max_steering, mass |
| $P_\phi(b_{t+1}\|s_t)$ | Background-vehicle behavior distribution | acceleration, lateral_acceleration, jerk, min_ttc, min_thw, min_headway |

*Table A.2. Training Scenario Set Sampling*

**Algorithm 1. Training Scenario Set Sampling**

**Require:** scenario pool $P = \{s_i\}_{i=1}^{M}$, target distribution $\phi$, number of scenarios $N$, feature set $F$, tolerances $\varepsilon$, swap budget $K$.
**Ensure:** sampled scenario set $S_\phi$, where $|S_\phi| = N$.

**for** each scenario $s_i$ in $P$ **do**
  $z_i \leftarrow ExtractFeatures(s_i, F)$
  **if** $z_i$ violates GeneLock **then** remove $s_i$
  **end if**
**end for**
$S \leftarrow UniformSample(P, N)$ // seeded initialization
$\mu(S) \leftarrow \frac{1}{|S|}\sum_{s_i \in S} z_i$
$L^* \leftarrow D(S,\phi) = \sum_{m \in F}\left(\frac{\mu_m(S)-\phi_m}{\max(\varepsilon_m,\delta)}\right)^2$
$S^* \leftarrow S$
**for** $k = 1$ to $K$ **do**
  select randomly $s_{out}$ from $S$
  select randomly $s_{in}$ from $P \setminus S$
  $S' \leftarrow S \setminus \{s_{out}\} \cup \{s_{in}\}$ // one-swap proposal
  $\mu(S') \leftarrow \frac{1}{|S'|}\sum_{s_i \in S'} z_i$
  $L' \leftarrow D(S',\phi)$
  **if** $L' \le L^*$ **then**
    $S \leftarrow S'$; $S^* \leftarrow S'$; $L^* \leftarrow L'$
  **end if**
**end for**
materialize $S^*$ by hard-linking selected scenario files
**return** $S^*$

*Table A.3. Evaluation Function*

| Metric | | | Weight | |
|---|---|---|---|---|
| Waymo | Success Rate | $M_1$ | $\beta_1$ | 0.36 |
| | Collision Rate | $M_2$ | $\beta_2$ | -0.15 |
| | Out of Road | $M_3$ | $\beta_3$ | -0.09 |
| nuScenes | Success Rate | $M_4$ | $\beta_4$ | 0.24 |
| | Collision Rate | $M_5$ | $\beta_5$ | -0.10 |
| | Out of Road | $M_6$ | $\beta_6$ | -0.06 |
| Constant Term | | | $\beta_0$ | 40 |

$$F = \beta_0 + \sum \beta_i \cdot M_i$$

*Table A.4. Algorithm Parameters*

| Parameter | Setting |
|---|---|
| Evolutionary generations | 12 |
| Population size | 10 |
| Training steps per stage | 25k |
| Clone : crossover : mutation | 4: 3: 3 |
| Selected parents (top : random) | 3: 1 |
| Elite gene pool size | Top 20 |
| Waymo evaluation weight | 0.6 |
| nuScenes evaluation weight | 0.4 |

*Table A.5. Experimental Parameters*

| Parameter | Setting |
|---|---|
| Simulator | MetaDrive 0.4.3 |
| Environment type | ScenarioEnv |
| Number of LiDAR rays | 30 |
| Observation dimension | 112 |
| Action dimension | 2 |
| Training model | TD3 |
| Learning rate | 3e-4 |
| Batch size | 256 |
| Actor network | MLP(112, 256, 256, 2) |
| Critic network | MLP(114, 256, 256, 1) |
| Training steps | 300k |
| Evaluation interval | 25k |
| Common Waymo source scenarios | 400 |
| TrafficGen variants derived from Waymo sources | 1510 |
| CAT variants derived from Waymo sources | 658 |
| fitness evaluation set: Waymo | 50 |
| fitness evaluation set: held-out nuScenes | 50 |
| Final test set (Waymo) | 100 |
| Final test set (nuScenes) | 83 |
| Repeated runs | 5 |
| Evolutionary search seed | 42 |

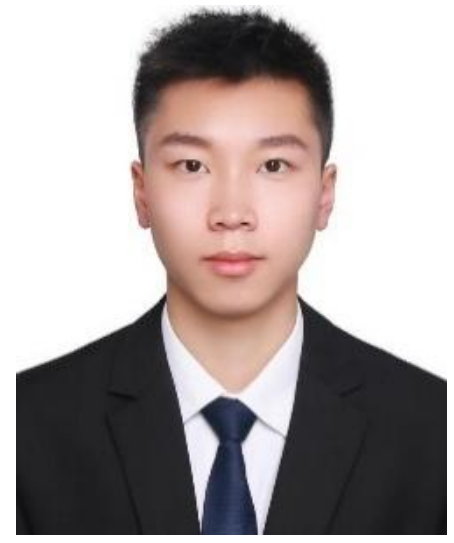

**Yichun Ye** received the B.S. degree in transportation engineering from Southwest Jiaotong University, Chengdu, Sichuan, China. He is currently working toward the Ph.D. degree with the Department of Traffic Engineering, Tongji University, Shanghai, China. His research focuses on safety testing of highly automated vehicles.

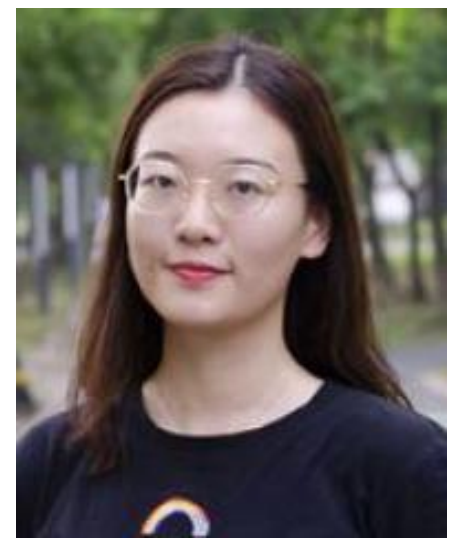

**He Zhang** received the Ph.D. degree in transportation engineering from Tongji University, Shanghai, China. She is currently a lecturer in transportation engineering at Southwest Jiaotong University. Her research focuses on safety testing of highly automated vehicles.

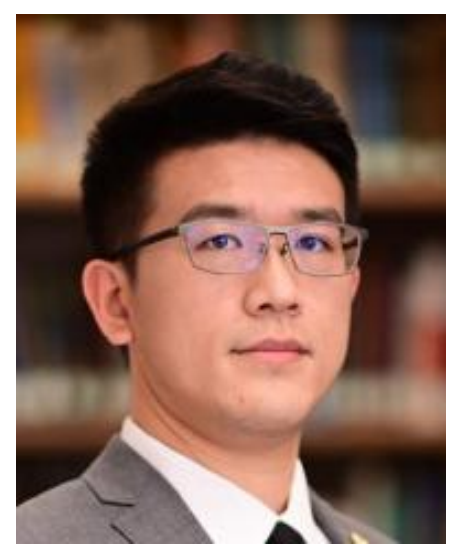

**Ye Tian** (Senior Member, IEEE) received the Ph.D. degree in transportation engineering from The University of Arizona, Tucson, AZ, USA, in 2015. He is currently a Professor of transportation engineering with Tongji University, Shanghai, China. He is an Associate Editor for IEEE TRANSACTIONS ON INTELLIGENT TRANSPORTATION SYSTEMS and IEEE TRANSACTIONS ON INTELLIGENT VEHICLES. His research interests include AI safety assurance and autonomous driving testing.

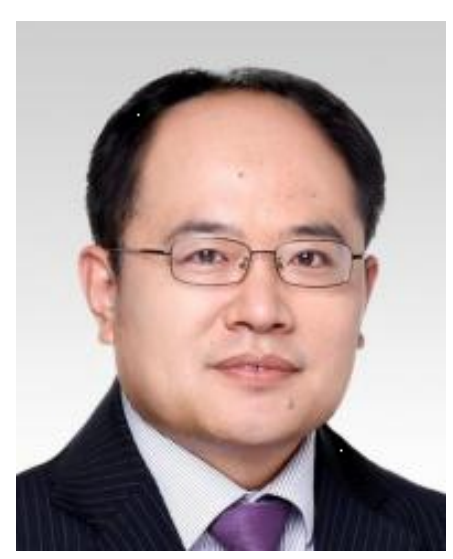

**Jian Sun** (Senior Member, IEEE) received the Ph.D. degree in transportation engineering from Tongji University, Shanghai, China. He is currently a Professor of transportation engineering with Tongji University. His research interests include intelligent transportation systems, traffic flow theory, AI in transportation, and traffic simulation.